\documentclass[sigconf,nonacm]{acmart} 
\usepackage[inline]{enumitem}

\AtBeginDocument{%
  }

\begin{document}

\title{Explainable Recommendations at Scale: LLM Rationales for YouTube Music Artist Discovery}

\author{Xiao Liu}
\email{liuxia@google.com}
\affiliation{%
  \institution{Google LLC}
  \country{USA}
}

\author{Yanwei Song}
\email{yanweisong@google.com}
\affiliation{%
  \institution{Google LLC}
  \country{USA}
}

\author{Srivaths Ranganathan}
\email{srivaths@google.com}
\affiliation{%
  \institution{Google LLC}
  \country{USA}
}

\author{Yuan Chen}
\email{ycyuanchen@google.com}
\affiliation{%
  \institution{Google LLC}
  \country{USA}
}

\author{Zheyun Feng}
\email{zfeng@google.com}
\affiliation{%
  \institution{Google LLC}
  \country{USA}
}

\author{Parker Steenburgh}
\email{psteenburgh@google.com}
\affiliation{%
  \institution{Google LLC}
  \country{USA}
}

\author{Jochen Klingenhoefer}
\email{jochenk@google.com}
\affiliation{%
  \institution{Google LLC}
  \country{USA}
}

\author{Nathan Lasche}
\email{lasche@google.com}
\affiliation{%
  \institution{Google LLC}
  \country{USA}
}

\author{Gergo Varady}
\email{gvarady@google.com}
\affiliation{%
  \institution{Google LLC}
  \country{USA}
}

\author{Tim Steele}
\email{timsteele@google.com}
\affiliation{%
  \institution{Google LLC}
  \country{USA}
}

\renewcommand{\shortauthors}{Liu, et al.}

\begin{abstract}
Modern music streaming platforms face a persistent tradeoff: exploiting familiar content versus driving the exploration of novel items. While users frequently desire discovery, they hesitate to select unknown artists over proven favorites. Providing transparent, natural language rationales that explain \textit{why} an unexplored item is recommended lowers this barrier. However, while Large Language Models (LLMs) excel at this nuanced explainability, their real-time deployment is severely bottlenecked by prohibitive inference costs and computational overhead. In this paper, we present an industry case study of a decoupled recommendation architecture that successfully scales exploration without compromising latency. Our system isolates LLM inference asynchronously offline, pre-computing personalized candidate pools of undiscovered artists alongside tailored rationales. Large-scale online A/B experiments validate our design. We demonstrate that combining LLM-backed recommendations with these explanatory rationales significantly reduces the trust barrier for new content, yielding statistically significant improvements in both user exploration and overall engagement on the discovery surfaces.

\end{abstract}



\maketitle

\section{Introduction}
Modern recommender systems increasingly rely on large, complex models to deliver highly personalized user experiences. Music streaming platforms, in particular, face two competing objectives: enabling users to quickly access their favorite songs for the current moment, while simultaneously helping them organically discover novel music to sustain long-term engagement. Although users crave discovery, there is high barrier to pick an unfamiliar artist over a familiar one. We argue that the primary bottleneck to exploration is not a lack of relevant candidates, but a lack of \textit{confidence}. Providing transparent, natural language rationales, explaining exactly why an unfamiliar artist is recommended, acts as a cognitive bridge, significantly lowering this trust barrier for novel content.

Large Language Models (LLMs) offer unprecedented capabilities to operationalize this explainability \cite{zhang2024large}. While traditional collaborative filtering (CF) models yield opaque affinity scores \cite{zhang2020explainable} and frequently face challenges with data sparsity \cite{zhang2025coldstart}, LLMs leverage vast world knowledge to derive rich semantic connections from minimal interaction data. More importantly, they enable a paradigm shift towards \textit{generative explainability}, providing nuanced, human-readable rationales alongside zero-shot candidate generation \cite{zenml2025spotify}. Furthermore, by flexibly integrating rich personal context, LLMs unlock a profound degree of personalization that makes their reasoning uniquely appealing and relevant to the user.

Despite these transformative capabilities, integrating LLMs directly into production recommender systems presents severe operational bottlenecks. Generating rich, personalized rationales and candidate pools online for billions of users is impractical. The prohibitive computational expense and strict sub-second latency constraints of real-time inference prevent the direct deployment of LLMs for live user sessions \cite{raja2025comprehensive}.

To bridge the gap between generative explainability and strict production constraints, we propose an industry-scale, decoupled recommendation architecture using Gemini. Our system shifts the heavy computational burden of LLM reasoning to an asynchronous offline environment. In this offline phase, the system analyzes a user's comprehensive listening history to compute personalized "discovery profiles" containing novel artist recommendations paired with compelling, natural language rationales. To ensure industry-level reliability and systematically eliminate hallucinations, this pipeline is fortified by an automated LLM-as-a-Judge evaluation framework \cite{zheng2023judging} and Knowledge Graph (KG) canonicalization that maps generated content to real-world entities.\cite{niu2024mitigating}. During the live user session, these robust profiles are served via a low-latency retrieval and annotation layer, entirely circumventing real-time inference costs.

We validated our explainable architecture through a live, large-scale A/B experiment on the YouTube Music platform. The results demonstrate that coupling LLM-backed recommendations with tailored rationales significantly lowers the trust barrier for new content, allowing novel candidates to bypass rigid heuristic similarity thresholds. The decoupled system yielded a statistically significant +22.43\% lift in user engagement on our discovery surfaces with negligible latency, expanded the pool of eligible discovery candidates, and successfully maintained strict sub-second response times.

\begin{figure}[htbp]
  \centering
  \includegraphics[width=0.3\textwidth]{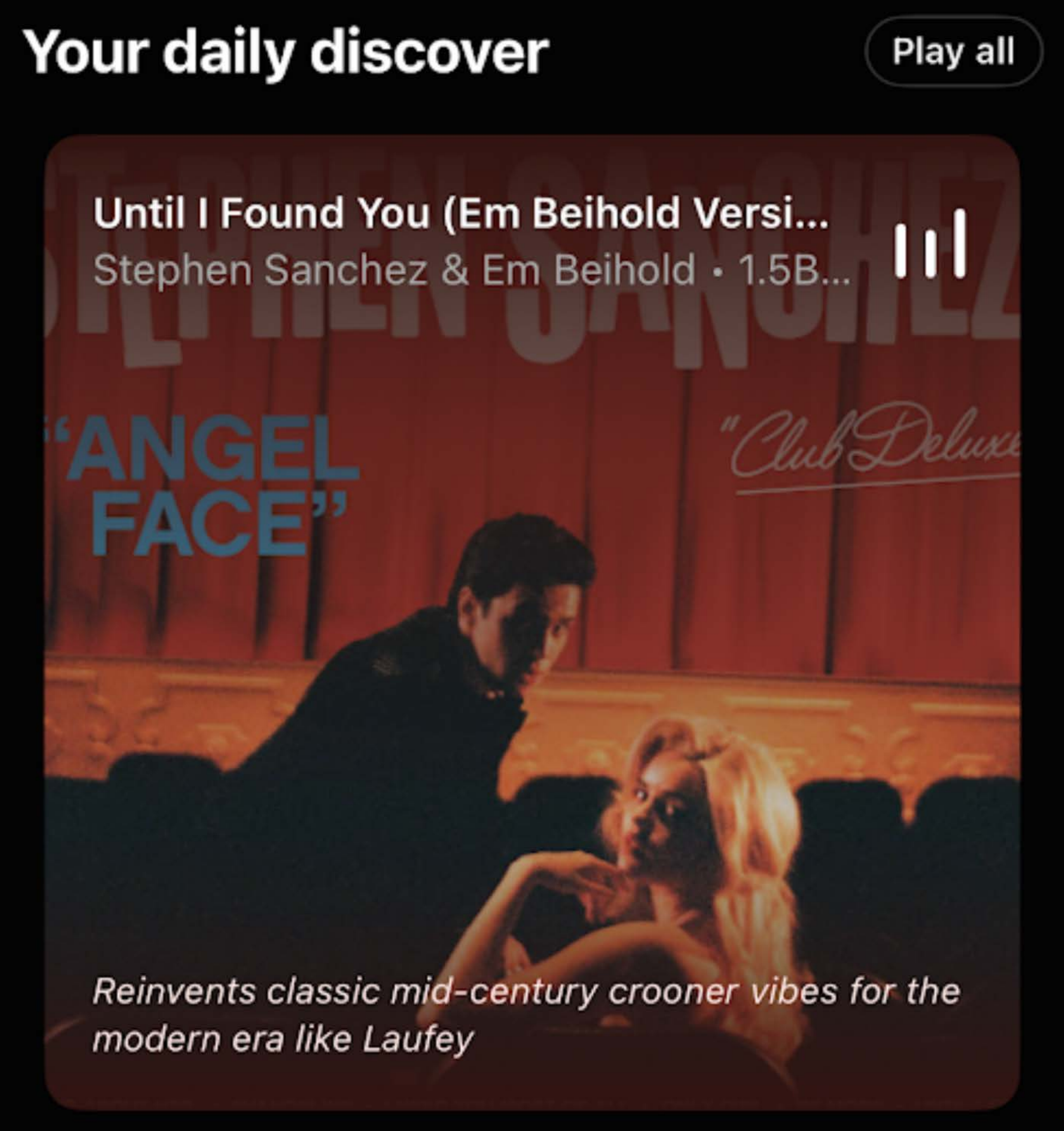}
  \caption{An example of the "Your daily discover" shelf in the YouTube Music app. The italicized text at the bottom provides a personalized rationale for the recommended track.}
  \label{fig:ydd}
  \Description{A screenshot of the YouTube Music app showing a recommended song card in the "Your daily discover" shelf. The card features the song "Until I Found You" by Stephen Sanchez and Em Beihold. At the bottom of the card, an italicized personalized rationale reads: "Reinvents classic mid-century crooner vibes for the modern era like Laufey".}
\end{figure}

\begin{figure*}[t]
  \centering
\includegraphics[width=0.7\textwidth]{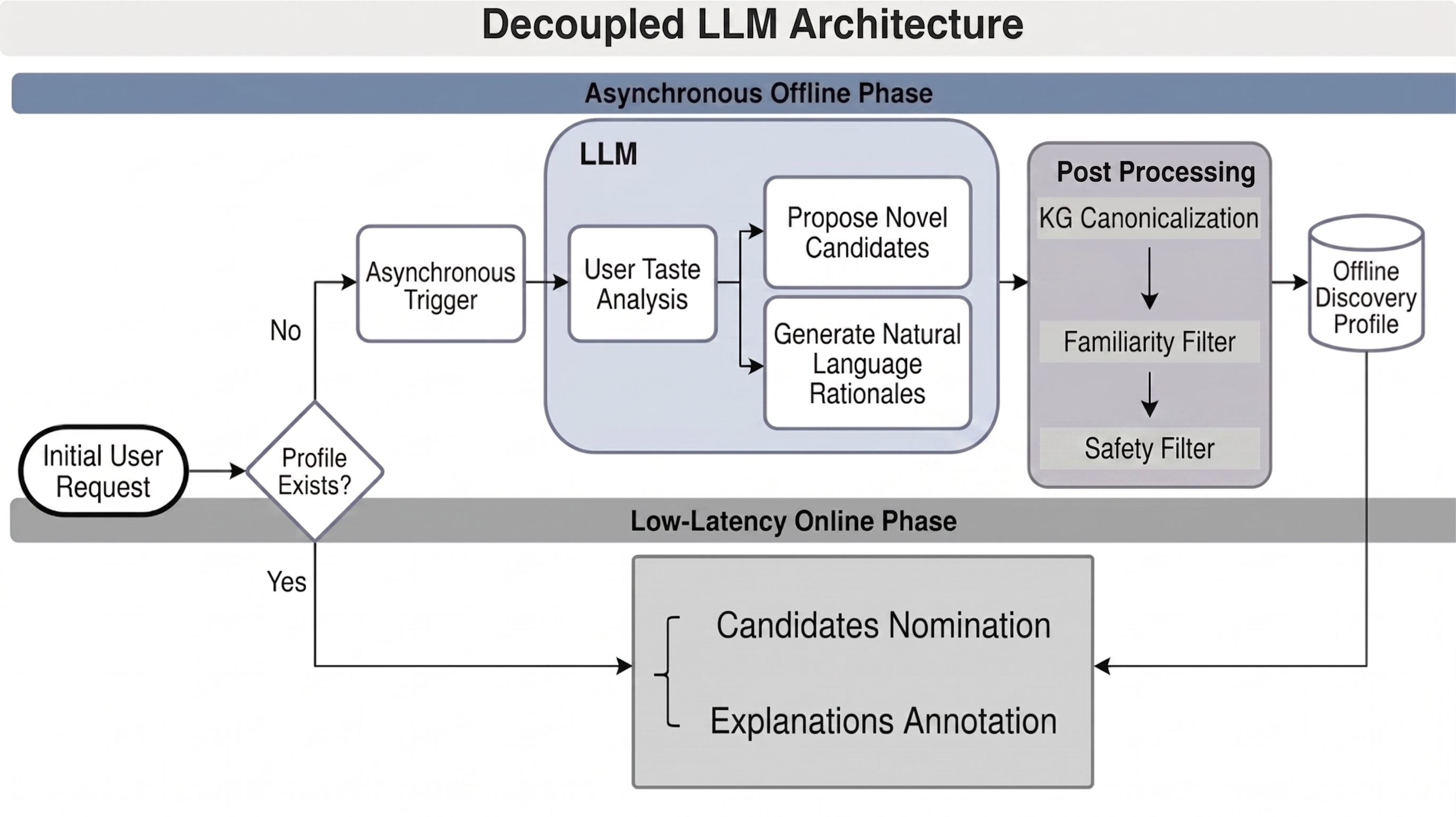}
  \caption{Overview of the decoupled two-phase recommendation architecture. An offline LLM asynchronously generates personalized discovery profiles containing novel candidates and rationales, which a low-latency online layer retrieves during live sessions.}
  \label{fig:llm_arch}
  \Description{Diagram of the Decoupled LLM Architecture showing Asynchronous Offline Phase and Low-Latency Online Phase.}
\end{figure*}

In summary, the main contributions of this paper are as follows:
\begin{itemize}
    \item \textbf{Decoupled Explainable Architecture:} We present a two-phase recommendation system, successfully launched in YouTube Music as illustrated in Figure~\ref{fig:llm_arch}, that decouples expensive LLM reasoning from live serving. This design enables the use of foundational models for generative explainability while strictly adhering to sub-second online latency constraints.
    \item \textbf{Automated Quality and Alignment Framework:} We propose a scalable offline evaluation strategy utilizing an LLM-as-a-Judge. Coupled with Knowledge Graph canonicalization, this framework systematically filters hallucinations, guarantees recommendation novelty, and drives iterative prompt tuning without requiring costly human annotation.
    \item \textbf{Large-Scale Empirical Validation:} We validate our approach through a live A/B experiment on YouTube Music. Our results provide industry-scale evidence that natural language rationales act as a cognitive bridge, successfully lowering the user trust barrier for unfamiliar artists and significantly lifting exploration and overall engagement.
\end{itemize}

The remainder of this paper is organized as follows. Section \ref{sec:related_work} reviews relevant prior work in industrial explainable recommendation, music recommendation, and LLM frameworks. Section \ref{sec:method} details our decoupled two-phase architecture, encompassing the asynchronous offline profile generation, our LLM-as-a-Judge evaluation framework, and the online serving mechanisms. Section \ref{sec:experiment} presents the experimental setup and the main results of our large-scale online A/B test on YouTube Music. Finally, Section \ref{sec:conclusion} concludes the paper with a discussion of our findings.

\section{Related Work}
\label{sec:related_work}
Our framework builds upon recent advancements in Large Language Models for recommendation, specifically focusing on conversational explainability, domain-specific music generation and industrial scalability.

\textbf{Conversational \& Explainable Recommenders} Industry applications, such as Spotify’s "AI DJ," have shown that providing human-readable context significantly boosts user confidence \cite{zenml2025spotify}. Recent academic frameworks have formalized this: Chat-REC \cite{gao2023chat} utilizes real-time LLM prompting to provide interactivity and explainability, while ReFICR \cite{yang2024unleashing} uses retrieval-augmented generation (RAG) to handle evolving item corpuses in conversational settings. While real-time prompting provides high interactivity, it is difficult to scale under strict latency budgets. To ensure reliability at scale, our system eschews real-time generation in favor of pre-computed rationale pools. We utilize an offline "LLM-as-a-Judge" alignment strategy and a Knowledge Graph (KG) canonicalization pipeline to enforce strict quality, filter out hallucinations, and guarantee novelty before any explanations reach the live serving layer.

\textbf{Generative Recommender Systems} Recent advancements in recommender systems have increasingly explored generative retrieval, transitioning from traditional multi-stage pipelines to cohesive sequence generation tasks. Pioneering frameworks such as TIGER \cite{rajput2023recommender} reframe candidate retrieval by autoregressively decoding discrete, semantically meaningful item identifiers (Semantic IDs) directly from user interaction histories. Extending this paradigm, unified models like the OneRec series \cite{deng2025onerec}\cite{zhou2025onerec} streamline the entire recommendation cascade, compressing retrieval, ranking, and optimization into a single generative architecture refined through iterative preference alignment. While these approaches successfully leverage generative pathways to directly output candidate items, their primary focus remains on structural pipeline unification and prediction efficiency. In contrast, our framework distinctively utilizes LLMs to produce personalized, natural language rationales that explain recommendations directly to the user. Rather than focusing on retrieval mechanics, our effort shifts the generative paradigm toward explainability, using semantic insights to lower the user trust barrier during content discovery.

\textbf{Music-Specific LLM Recommendation} The unique challenges of music discovery—such as nuanced cultural contexts, audio feature synthesis, and abstract user queries—make the application of LLMs particularly complex. Concurrent works have attempted to address this through various mechanisms. For instance, TALKPLAY \cite{doh2025talkplay} treats music recommendation as an end-to-end token generation problem using a multimodal tokenizer. In the broader domain of sequential recommendation, LLM2Rec \cite{he2025llm2rec} demonstrates that LLMs can be adapted into generalizable embedding models by explicitly integrating CF awareness through collaborative supervised fine-tuning. In contrast to these approaches, which often rely on expensive domain-specific fine-tuning or end-to-end generation, our work leverages the zero-shot reasoning capabilities of Gemini. By shifting optimization to prompt tuning and structured Chain-of-Thought reasoning \cite{wei2022chain}, we generate novel content directly from foundational listening identities without requiring structural model alterations.

\textbf{Offline/Online Decoupled \& Industrial LLM Frameworks} Recent research has increasingly explored paradigms to embed LLM capabilities into recommendation architectures while mitigating real-time inference costs \cite{raja2025comprehensive, bao2023tallrec, ranganathan2026multi}. To circumvent latency bottlenecks, frameworks such as ReLand \cite{tian2024reland} propose leveraging an "LLM Reasoning Pool" by performing generative recommendations offline on seed users, and subsequently using retrieval to attach reliable recommendation rationales online for the broader user base. Similarly, systems like HyGen address the latency-throughput dichotomy by efficiently co-locating offline throughput-oriented LLM workloads with online tasks. Our approach extends these paradigms by utilizing opportunistic TPU \cite{jouppi2017datacenter} batch processing and configurable refresh cycles to generate personalized discovery profiles asynchronously, balancing cost efficiency with content freshness for tens of millions of daily active users.

\section{Method and System Design}
\label{sec:method}
Real-time recommendation architectures are inherently constrained by strict latency budgets. To meet these demands, conventional systems are typically forced to process truncated interaction sequences, which frequently causes models to overfit to transient engagement spikes. This yields disjointed, overly reactive discovery sessions. Conversely, while Large Language Models (LLMs) offer the advanced semantic reasoning required to synthesize a user's comprehensive listening history, their immense computational overhead and inference latency preclude direct deployment in these real-time environments.

To resolve this dichotomy, we implemented a decoupled, two-phase architecture, as illustrated in Figure~\ref{fig:llm_arch}, that isolates the computationally intensive recommendation and reasoning processes from the latency-sensitive online serving layer. By shifting profile generation to an asynchronous offline environment, our framework completely neutralizes real-time LLM inference latency. Furthermore, removing these temporal constraints allows the system to bypass forced data truncation, enabling the safe ingestion of a user's complete, long-term listening history.

Equipped with this expansive context, the LLM analyzes the user's complete listening history to distinguish between short-term phases and deeply rooted sonic preferences. By grouping these insights into distinct taste clusters, the system ensures every generated rationale is anchored directly to the user's core musical identity. Consequently, it can transparently contextualize novel recommendations, using these natural language explanations as a cognitive bridge to lower the trust barrier for unfamiliar artists.

\subsection{Decoupled Profile Generation}
Unlike traditional batch-processing architectures that proactively compute recommendations across an entire user base, our system employs an event-driven, asynchronous trigger. A profile generation request is dispatched strictly on-demand---only when an eligible user actively accesses the YouTube Music Homepage. To guarantee zero degradation to the initial user experience, the online serving layer seamlessly falls back to standard, heuristic-based discovery if a pre-computed profile is not yet available. Once the offline LLM inference concludes, the resulting \textit{Discovery Profile} is securely cached and immediately available to enrich subsequent user sessions. This lazy-generation strategy strictly limits expensive inference operations to active users, drastically reducing unnecessary computational overhead and preserving system efficiency at scale.

\subsection{LLM-based Candidate and Rationale Generation}
At the core of our generative pipeline, the LLM transforms a user's raw music engagement data into structured discovery clusters. Specifically, the model analyzes the comprehensive listening history to segment the user's preferences into distinct, cohesive taste groups. For each identified group, the system proposes novel candidate artists that are semantically aligned with the cluster but are likely undiscovered by the user. Concurrently, the LLM generates natural language rationales that explicitly articulate the clear connection between the user's established tastes and these new recommendations. 

\begin{figure}[ht]
  \centering
  \includegraphics[width=\linewidth ]{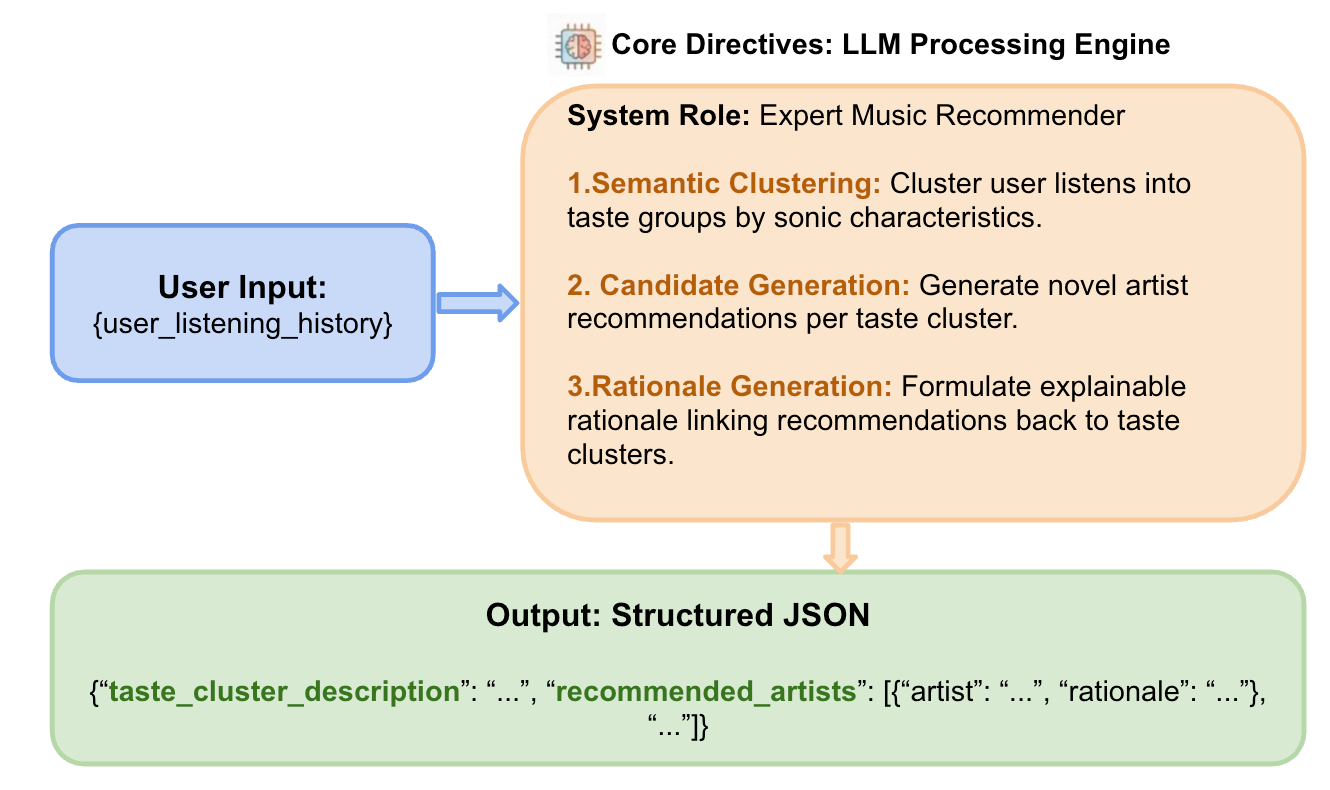}
  \caption{Prompt workflow for candidate and rationale generation. The LLM segments a user's comprehensive listening history into cohesive taste clusters, utilizing these clusters as a foundation to nominate novel artists and generate personalized rationales.}
  \label{fig:generation_prompt}
  \Description{Diagram of Candidate Generation Prompt with User Input and LLM Processing Engine.}
\end{figure}

As detailed by the prompt structure in Figure~\ref{fig:generation_prompt}, this workflow inherently acts as a Chain-of-Thought (CoT) reasoning process. By requiring the model to generate the rationale alongside the nomination, this approach simultaneously enhances the semantic quality of the candidate recommendations and yields the user-facing explanations required for the live application.

\subsection{Offline Evaluation and Prompt Tuning via LLM-as-a-Judge}
Deploying generative LLMs in production recommender systems requires rigorous quality guardrails. To evaluate recommendation quality at scale without relying on prohibitively expensive human annotations, we employ an ``LLM-as-a-Judge'' alignment strategy \cite{zheng2023judging}. We utilize a highly capable offline reasoning model (Gemini 3.1 Pro) \cite{team2023gemini} to act as an automated rater. This judge systematically evaluates both the structural cohesion of the generated taste groups and the semantic alignment of the novel artists proposed by the candidate generator.

\begin{figure}[ht]
  \centering
  \includegraphics[width=\linewidth]{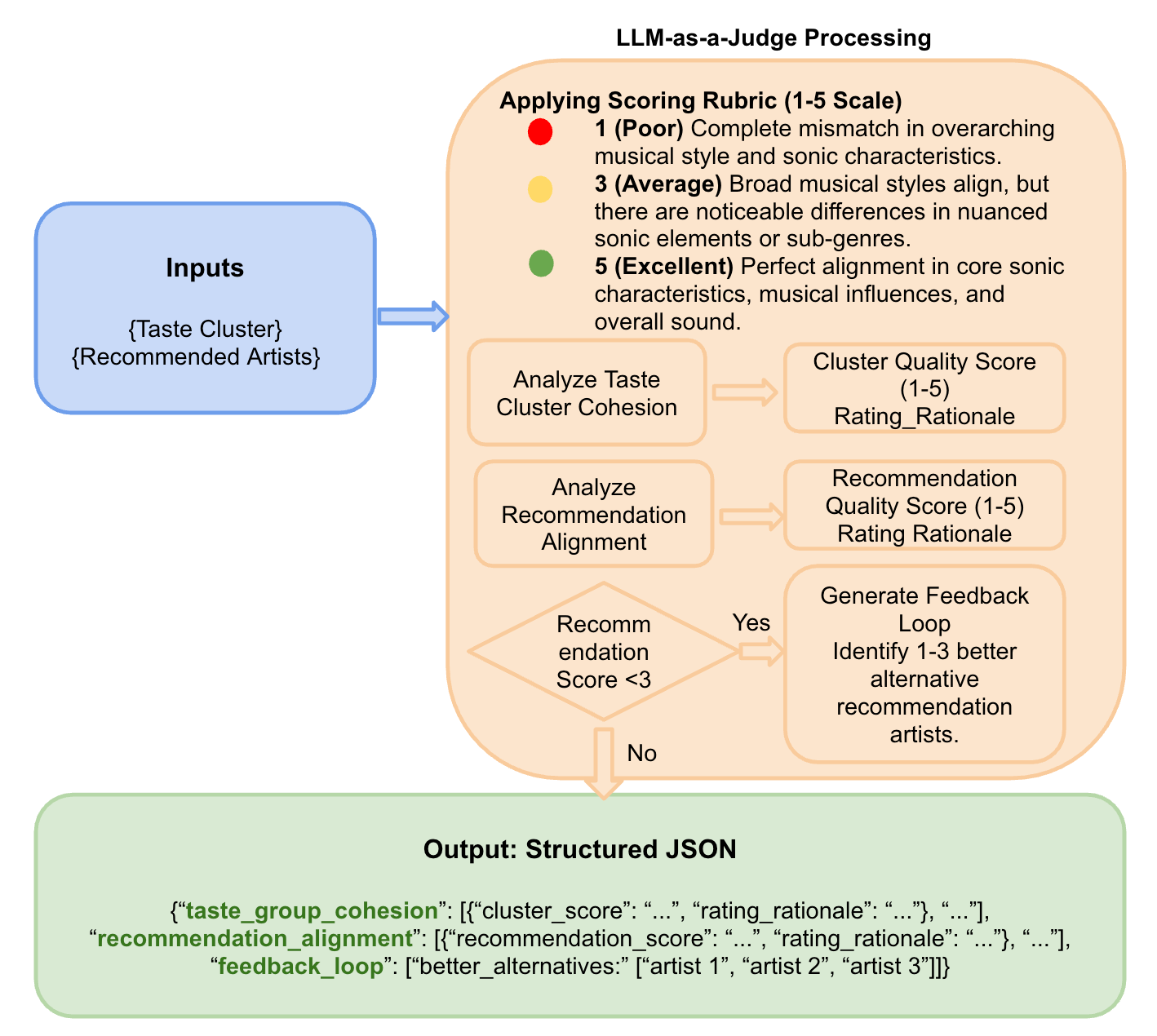}
  \caption{A multi-step LLM-as-a-Judge evaluation rubric detailing three phases: assessing taste cluster cohesion, scoring candidate semantic alignment, and a targeted feedback loop that triggers on low scores to provide alternatives and drive prompt tuning.}
  \label{fig:llm_judge_prompt}
  \Description{Diagram of LLM-as-a-Judge processing.}
\end{figure}

As illustrated in Figure~\ref{fig:llm_judge_prompt}, the evaluation process is structured as a sequential, multi-step rubric. By decoupling the assessment into distinct analytical phases, the framework precisely isolates the origin of qualitative degradation, differentiating between structural failures in foundational taste group generation and semantic misalignments in downstream candidate nomination:
\begin{itemize}
    \item \textbf{Taste Cluster Cohesion Assessment:} The judge initially evaluates the internal consistency of the generated taste group. It conducts a multi-dimensional analysis of the cluster's uniformity across primary subgenre, vocal styling, cultural context, and linguistic attributes. By scoring the cluster's consistency on a 1-to-5 scale, the system can determine whether poor recommendations are simply the result of a poorly formed taste group.
    \item \textbf{Recommendation Alignment Scoring:} Subsequently, the framework quantifies the semantic alignment between the proposed novel artists and the established taste cluster. Rated on the same 1-to-5 scale, this phase targets candidate generation efficacy, measuring the precise likelihood that a user anchored to the seed cluster will positively engage with the recommended artist.
    \item \textbf{Targeted Feedback Loop:} Finally, any recommendation receiving a score below 3 triggers an automated feedback mechanism. The judge utilizes forward-reasoning to output a detailed rationale for the deficiency and identifies 1 to 3 alternative artists demonstrating superior fit. This step facilitates the systematic aggregation of recurring loss patterns, directly informing iterative prompt tuning to resolve localized model failures.
\end{itemize}

\subsubsection*{Iterative Prompt Tuning via Loss Pattern Analysis}

Recommendations scoring below 3 are systematically flagged as negative examples. By aggregating these automated critiques, the judge model served as a scalable engine to identify recurring qualitative ``Loss Patterns'' in the generator's behavior. The primary failure modes identified during our evaluations included:
\begin{enumerate}
    \item \textbf{Semantic Mismatches:} Cases where the recommended artist's music style or genre materially deviated from the cluster description.
    \item \textbf{Entity Hallucination:} Occurrences of the model fabricating rationales or recommending non-musicians, such as a film director, lyricist, or screenwriter.
    \item \textbf{Linguistic Inconsistencies:} Language mismatches within the proposed cluster.
\end{enumerate}

This automated identification of loss patterns formed the foundation of our prompt tuning strategy. Rather than pursuing expensive and brittle fine-tuning of model weights, we iteratively refined the generation prompt to explicitly guard against these failure modes. For example, detecting language mismatches prompted us to inject strict prioritization constraints into the system (e.g., instructing the model to segment all artists by their primary language before conducting any further stylistic analysis).

Empirically, this prompt engineering loop yielded substantial gains. Prior to adjustments based on the loss pattern analysis, our baseline serving prompt achieved a passing recommendation quality score ( $> 3$ ) on only 64.8\% of the sampled clusters. By integrating the structural rules derived from the LLM judge's feedback, the refined prompt improved the absolute acceptance rate to 74.4\%. This demonstrates that an LLM-based offline evaluation framework can serve as a highly effective, scalable engine for iterative system alignment prior to live online A/B testing.

\subsection{Post-Processing}
To systematically mitigate Large Language Model (LLM) hallucinations and enforce recommendation novelty, we implemented a robust, two-stage post-processing architecture. 

The first stage acts as a deterministic canonicalization layer, resolving generated artist strings against a canonical Knowledge Graph (KG). During the generation phase, we empirically observed the LLM producing morphologically sound but entirely non-existent entities. These ranged from plausible Western nomenclatures (e.g., `Gavin DeBardeleben') to seamlessly combined, authentic Khmer syllables (e.g., `Chhet Sokun' or `Touch Sokhen'). Because these combinatorial fabrications are contextually and linguistically indistinguishable from legitimate, long-tail artists, traditional heuristic or text-based filtering methods prove fundamentally inadequate. By mandating that every recommended candidate accurately maps to a verifiable KG entity ID, this mechanism deterministically prunes hallucinated artists.

Subsequently, a familiarity filter cross-references these grounded entities against the user's comprehensive listening history. Crucially, by operating within our decoupled offline framework, this filtering layer can afford to ingest and scan the user's \textit{complete} historical engagement, avoiding the data truncation typically required in latency-sensitive online serving. This secondary layer strictly excludes any previously consumed artists, ensuring that the final candidate pool delivers genuinely undiscovered content and fulfills the core objective of the discovery shelf.

Finally, the pipeline integrates dedicated safety filters to detect and prevent policy-violating content. During this validation phase, any non-compliant LLM rationales are systematically discarded. This strict enforcement layer guarantees that all surfaced natural language explanations remain safe and fully aligned with platform community guidelines.

\subsection{Online Nomination and Annotation}
The online serving layer is designed to seamlessly integrate these pre-computed LLM insights without impacting real-time latency SLAs. During a live user request, the system executes a low-latency retrieval of the user's cached \textit{Discovery Profile}. This profile serves a dual purpose: first, it nominates novel, KG-grounded artist and song candidates to supplement traditional collaborative filtering retrieval sources; second, it provides the corresponding pre-generated natural language rationales. By performing rationale annotation as a lightweight fetch operation rather than a live generative inference step, the system successfully enriches targeted discovery surfaces, such as the ``Your Daily Discovery''(YDD) shelf on the YouTube Music home page, with highly personalized, explainable recommendations while strictly maintaining sub-second response times.

\subsection{Efficiency and Cost Optimization}
Operating at the scale of tens of millions of daily active users necessitates rigorous resource optimization. We maximize system efficiency by fully exploiting the flexibility of our decoupled, asynchronous architecture. 
The primary cost-saving mechanism is opportunistic TPU utilization; because profile generation is not bound by user-facing latency constraints, inference workloads are scheduled during off-peak hours or routed to lower-priority, preemptible compute tiers. Furthermore, we implemented configurable refresh cycles to decouple profile staleness from daily traffic spikes. By strategically defining the update intervals for these profiles, we achieve an optimal equilibrium between content freshness and computational overhead. Ultimately, this architectural design ensures that serving state-of-the-art, LLM-driven discovery experiences remains economically viable at a global scale, entirely eliminating the reliance on dedicated, high-availability inference clusters.

\section{Experiment}
\label{sec:experiment}
We evaluated the proposed decoupled architecture via a large-scale online A/B test on the YouTube Music platform. The experiment targeted the YDD shelf, a high-traffic surface dedicated to surfacing novel content and expanding user music horizons.

\subsection{Baseline Configuration for YDD}\label{sect: ydd_config}
 The baseline setup for the YDD shelf follows the foundational two-stage architecture for YouTube recommendations \cite{covington2016deep}, consisting of components, in sequence, for  \begin{enumerate*} \item candidate retrieval, \item heuristic rationale generation, \item candidate ranking, and \item shelf packing. \end{enumerate*} The baseline candidate generation depends on a collection of retrieval models. For each user, this collection of models finds an initial set of personalized song and video recommendations. The initial set of recommendations serves different experiences on the YouTube Music Home page (i.e., shelves other than YDD) and may contain content that is familiar to the user. We select only the discovery-eligible candidates, those that do not appear in the user's watch history, from this initial set for further processing.

Existing heuristic rationales on YDD focus on similarities between the generated candidates and  personalized seed entities with which the user is highly familiar. Example heuristic rationales include ``For Fans of [Seed Artist]'', which indicates that users who listen to a specific seed artist typically also listen to the recommended song, and ``Sounds like [Seed Song]'', which indicates that the recommended song and seed song have similar audio properties. To generate these heuristic rationales, we retrieve a set of seed entities, both artists and songs, based on user's watch history, preferring entities with which the user has had higher and more recent levels of engagement. Then, we compute the pairwise cosine similarity distances between each seed entity and recommended candidate in several, pre-trained embedding spaces. These embeddings semantically correspond the different rationale categories, e.g., we use a co-watch based embedding for the ``For Fans of [Seed Artist]'' rationale, and we use an audio waveform based embedding for the ``Sounds like [Seed Song]'' rationale. 

For each recommended candidate and for each embedding type, we find the \textit{most similar} seed entity. If the similarity measure exceeds a given threshold, then we annotate the recommended candidate with the corresponding rationale. I.e., we annotate a recommended candidate with the ``For Fans of [Seed Artist]'' rationale if and only if there exists a seed artist with a high enough cosine similarity to the candidate in the co-watch based embedding space. In the end, we only select and pack candidates with attached rationales to the YDD shelf. This requirement ensures that the recommended candidates are relevant and similar to the user's music tastes.

After rationale computation, we score the recommended candidates using a deep ranking model \cite{ranganathan2025zero} that predicts several dimensions of engagement (click through rate, conditional watch time, discovery rate, etc.). An auxiliary model aggregates the engagement predictions into a single ``reward'' prediction for each candidate. We sort the candidates according to this reward and pack the top ranking candidates into the final YDD shelf, subject to business logic diversity constraints (e.g., to ensure that there is diversity in the rationale categories shown to the user).

\subsection{Experimental Setup}
We conducted a two-week A/B test to evaluate stable user behavior, partitioning users into two cohorts:
\begin{itemize}
    \item \textbf{Control}: Received the production experience described in Section~\ref{sect: ydd_config}, where rationales for discovery candidates depend rule-based heuristic templates (e.g., "For Fans of [Artist]" or "Sounds Like [Song]").
    \item \textbf{Treatment}: Received the integrated LLM experience, detailed below, which injected novel candidates from the pre-computed discovery profiles and applied rich, natural-language rationales to all eligible items.
\end{itemize}

In the treatment cohort, we augment recommendations of the baseline YDD configuration with candidates and rationales from the user's discovery profile. We apply two broad augmentation methods to treatment cohort: \begin{enumerate*} \item directly adding candidates to the recommendation set, and \item attaching LLM rationales to \textit{existing} candidates. \end{enumerate*} The discovery profile contains relevant \textit{artists} for the user to discover. 

To directly add \textit{song and video} candidates to the recommendation set, we leverage restricted nomination capabilities of existing retrieval models in the baseline configuration. Specifically, our existing retrieval models are capable of recommending candidates with attribute constraints. In this case, we supplement the generic recommendations with recommendations of songs and videos by the artists that appear in the user's discovery profile. Aside from directly expanding the recommendation set with new candidates, we also use attach the LLM-generated rationales in the discovery profile to \textit{all} eligible candidates in the recommendation set. This includes recommendations from the baseline YDD configuration (it is possible for the baseline YDD configuration to coincidentally recommend candidates by artists that happen to be in the user's discovery profile). 

\subsection{Main Results}
Integrating LLM-driven nomination and annotation yielded statistically significant gains across both engagement and discovery metrics. Notably, the new nominator proved highly effective at surfacing novel content and, after this was launched, the YDD shelf organically elevated into YouTube Music's top three positions at a significantly higher rate, reflecting a substantially improved discovery experience.

\begin{table}[ht]
  \caption{Impact of LLM Rationales on Key Discovery Metrics}
  \resizebox{\columnwidth}{!}{%
    \begin{tabular}{lcc}
      \toprule
      \textbf{Metric} & \textbf{Value} & \textbf{95\% Confidence Interval} \\
      \midrule
      New Nominator Views & 4.1\% & N/A \\
      New Nominator Uniqueness Rate & 74.00\% & N/A \\
      YDD Shelf-Level Engagement & +22.43\% & [16.93\%, 27.93\%] \\
      YDD Shelf-Level Discovery Metric (Retention) & +8.07\% & [0.90\%, 15.24\%] \\
      \bottomrule
    \end{tabular}%
  }
\end{table}

As detailed in Table 1, although direct consumption from the new nominator drove a modest 4.1\% increase in views, the aggregate YDD shelf-level engagement and discovery metrics exhibited substantial gains. To disentangle the relative contributions of LLM nomination and rationale annotation, we conducted a multi-arm holdback experiment. Comparing the full treatment holdback against individual feature holdback reveals that the LLM rationale annotation was the primary driver of the shelf's overall engagement gain. Specifically, disabling only the LLM rationale annotations, while keeping the visual interface and candidate pool unchanged, precipitated engagement regressions that closely mirrored the degradation observed when the entire LLM system was withheld. Conversely, disabling only the new nomination source resulted in a neutral engagement impact. Nevertheless, evaluating the discovery metric highlights a critical systemic interdependence: discovery performance degraded significantly under a full holdback, and experienced partial regressions when either feature was independently omitted.

These results empirically confirm the model's capacity to serve as a ``cognitive bridge.'' By utilizing natural language rationales, the system effectively lowers the user trust barrier typically associated with unfamiliar content. Furthermore, this generative mechanism liberates the discovery experience from the rigid similarity constraints inherent to traditional heuristic templates, thereby surfacing a significantly more diverse candidate pool. From an operational standpoint, the deployment definitively validated the efficacy of our decoupled architecture. Despite the integration of complex LLM-derived insights, the mean online service latency increased by a negligible 0.3\%, successfully preserving the strict sub-second response times and high availability essential for production-scale streaming.

\section{Conclusion}
\label{sec:conclusion}
This paper introduced a novel, two-phase recommendation framework for guided exploration that successfully bridges the gap between the advanced generative capabilities of Large Language Models and the strict latency constraints of production-scale streaming platforms. By isolating computationally intensive reasoning into an asynchronous offline phase, our system leverages the extensive world knowledge of LLMs to generate highly personalized artist nominations paired with natural language rationales. Furthermore, we demonstrated that the inherent risks of generative models can be systematically mitigated at scale through an automated LLM-as-a-Judge evaluation pipeline and deterministic Knowledge Graph canonicalization. Large-scale online A/B testing on YouTube Music validated the effectiveness of this architecture, driving significant gains in user engagement and discovery metrics. These results confirm that transparent rationales act as a powerful cognitive bridge, effectively lowering the trust barrier for novel content. Ultimately, this work provides a practical, robust blueprint for the industry to transition toward explainable, LLM-driven recommender systems without compromising the minimal latency demanded by real-time production environments.

\bibliographystyle{unsrtnat}
\bibliography{main} 

\end{document}